\documentclass[runningheads]{llncs}
\usepackage{graphicx}
\usepackage{mathtools}
\usepackage{times}
\usepackage{latexsym}
\usepackage{textcomp}
\usepackage{xcolor}
\usepackage{array,multirow}
\usepackage{float}
\usepackage{mathtools}
\usepackage{dirtytalk}
\usepackage{times}
\usepackage{latexsym}
\usepackage{breqn}
\usepackage{amssymb,amsfonts}
\usepackage{booktabs} 
\usepackage{amsmath}
\usepackage{bbm}
\usepackage{amssymb,amsfonts}
\usepackage[noend]{algpseudocode}
\usepackage{tabularx}
\usepackage[ruled,vlined]{algorithm2e}
\usepackage{lipsum}
\usepackage{algpseudocode}
\begin{document}
\title{Adaptive Summaries:\\ A Personalized Concept-based Summarization Approach by Learning from Users' Feedback}
\titlerunning{Adaptive Summaries}
\author{Samira Ghodratnama\inst{1} \and
Mehrdad Zakershahrak\inst{2}\and
Fariborz Sobhanmanesh\inst{1}}

\def\algbackskip{\hskip-\ALG@thistlm}
\makeatother

\authorrunning{S. Ghodratnama et al.}
\institute{Macquarie University, Sydney, Australia
\email{\{samira.ghodratnama,fariborz.sobhanmanesh\}@mq.edu.au} \and
Arizona State University, Arizona, United States\\
\email{\{mehrdad\}@asu.edu}}

\maketitle          
\begin{abstract}
\label{abs}
Exploring the tremendous amount of data efficiently to make a decision, similar to answering a complicated question, is challenging with many real-world application scenarios. 
In this context, automatic summarization has substantial importance as it will provide the foundation for big data analytic.
Traditional summarization approaches optimize the system to produce a short static summary that fits all users that do not consider the subjectivity aspect of summarization, i.e., what is deemed valuable for different users, making these approaches impractical in real-world use cases.
This paper proposes an interactive concept-based summarization model, called \textit{Adaptive Summaries}, that helps users make their desired summary instead of producing a single inflexible summary.
The system learns from users' provided information gradually while interacting with the system by giving feedback in an iterative loop.
Users can choose either reject or accept action for selecting a concept being included in the summary with the importance of that concept from users' perspectives and confidence level of their feedback.
The proposed approach can guarantee interactive speed to keep the user engaged in the process.
Furthermore, it eliminates the need for reference summaries, which is a challenging issue for summarization tasks.
Evaluations show that \textit{Adaptive Summaries} helps users make high-quality summaries based on their preferences by maximizing the user-desired content in the generated summaries.
\end{abstract}
\section{Introduction}
The expansion of Internet and Web applications, followed by the growing influence of smartphones on every aspect of our lives, has induced an everyday growth of textual information.
As a result, data summaries as a solution are becoming of paramount importance.
Therefore, carefully constructed summaries make the data analytic possible by improving scalability and efficiency.
Summarization has been widely used in many applications and domains, using a variety of techniques~\cite{amouzgar2018isheets,beheshti2018iprocess,schiliro2018icop,ghodratnama2021intelligent,summary2vec,zakershahrak2020we}.
A good summary should keep the main content while helping users understand large volumes of information in a small amount of time.
However, the summarization problem is subjective because different users have different attitudes toward what is considered valuable.
Consequently, producing a generic summary that can satisfy everyone makes the problem challenging.
Therefore, despite much research in this area, it is still a significant challenge to produce summaries that can satisfy all users.

Traditional state-of-the-art approaches produce only a single, globally short summary for all users~\cite{gupta2010survey}. 
They optimize a system towards one single best summary without considering users' interests and needs in seeking their desired information~\cite{avinesh2017joint}.
However, this is not useful in real-world scenarios where different users may explore diverse interests in the same corpus, thus need a distinct summary.
Furthermore, these high-level interests vary over time.
To be more specific, a person might be interested in a different area based on \textit{background knowledge}, and \textit{context} due to their cognitive bias.
For instance, there is various information available on the Internet about COVID-19.
While one might be interested in symptoms, the other could be looking for the outbreak locations, while others are searching about the death toll.
An example of background knowledge is when a researcher works on a  research topic, for instance, ``Summarization''.
She could be eager to know \textit{what is} the definition of summarization.
Then her interest may turn to different categories of summarization, such as extractive or abstractive approaches. 
Therefore, a good summary should change correspondingly based on the interest and preference of its reader. 

A recent definition of summarization is given by Radev et al.~\cite{radev2002introduction} as \say{a text that is produced from one or more texts, which conveys the important information in the original text, and usually significantly less than that.}
However, the \textit{importance}  interpretation in this definition is different even for one person according to the need, time, knowledge. 
Besides, humans quickly assess the importance of concepts from their side.
Previous approaches mainly select the most informative sentences as the summary and try to employ users' feedback in selecting sentences, not content, which makes the summaries vague. 
Therefore, it would be advantageous if users can interact with the system to incorporate their desired information into the summary.
While there exist many automatic summarization approaches, only a few methods focused on the needs of individuals.
Among them, a few considered the notion of the importance of a concept included in a summary, where this notion does not refer to users' attitudes and is statistical properties of the content, such as the frequency of occurrence of a word in a body of text~\cite{woodsend2012multiple}. 
Therefore, they fail to heed what is deemed to be valuable from individuals' perspectives.
One way to achieve personalized summary is thus by integrating the advantages of personal feedback in defining what is considered as important.

\begin{figure*}[t]
    \centerline{\includegraphics[width=\textwidth]{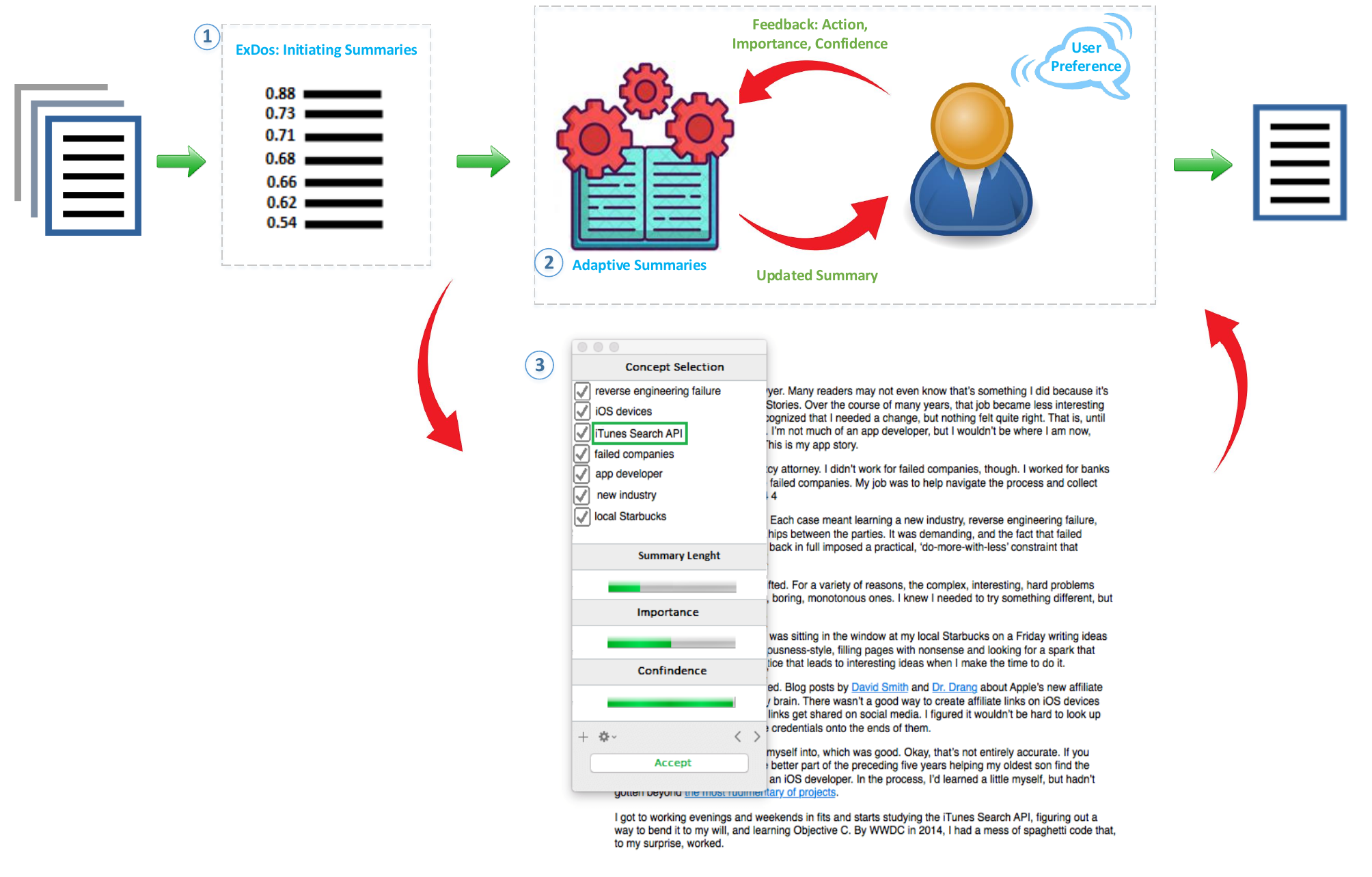}}
    \caption[]{An overview of the proposed approach (Adaptive Summaries). 1) Summaries are initiated with ExDos~\cite{samira2020}. 2) Users integrate their preferences in making summaries by giving feedback in an iterative loop. 3) An example of user interaction.}
    \label{overview}
\end{figure*}

We put the human in the loop and create a personalized summary that better captures the users' needs and their different notions of importance. Besides, the notion of having the human in the loop is very popular in different aspects of Explainable AI (XAI) ~\cite{zakershahrak2018interactive,zakershahrak2019online,zakershahrak2020order}.
In this setting, users can give feedback in an iterative loop in selecting or rejecting a concept, defining the level of importance or being unrelated, and giving the confidence level in their feedback.
By doing this, we allow even novice users to interactively explore, manipulate, and analyze sizeable unstructured text document collections to find their desired information and integrate their user-specific notion of importance.
Our model employs an integer linear programming (ILP) optimization function to maximize user-desired content selection.
Besides, most existing document summarization techniques require access to reference summaries to train their systems.
However, obtaining reference summaries is very expensive. 
Lin in~\cite{lin2004rouge} explains that 3,000 hours of human effort is required for a simple evaluation of the summaries for the Document Understanding Conference (DUC).
\textit{Adaptive Summaries} does not require reference summaries since it optimizes the summaries based on user-specific needs and not the goals standard summaries.
An overview of the proposed approach is illustrated in Figure~\ref{overview}. 
\textit{Adaptive Summaries} can be used in multiple application scenarios where there are numerous documents, and the users seek information for getting personal insights.
The main contributions can be summarized as follows:
\begin{itemize}
    \item We have proposed an algorithm, called \textit{Adaptive Summaries}, to include the user's needs and knowledge in making summaries. 
    Adaptive summaries help users select the content of summaries based on their perspective, defining the degree of importance, and confidence in their feedback which benefits users in various ways including:
    \begin{itemize}
        \item Customized Summary Length: The user can choose the length of the summary.
        \item Interaction: Users interact with the summary, and it provides a better understanding of the topic.
        Besides, interacting with summary hint users to understand what is important related to the documents.
        \item Reference Summary Requirements: The summaries' dynamic structure eliminates the needs for reference summaries since there is no need to optimize a summary based on reference summaries. 
    \end{itemize}
    \item We provide evidence in the form of simulated user-oriented experiments to prove the model helps users make their summaries.
    Adaptive summaries also have an additional advantage, which is being very interpretable.
\end{itemize}

The rest of this paper is organized as follows. Section 2 discusses state-of-the-art methods. Section 3 presents the proposed method and section 4 presents the experimental results. Section 5 discusses and justifies the obtained results. Finally, Section 6 concludes the paper.
\section{Related Work}
Producing a summary is quite a complicated task even for a person who has the domain knowledge of words and concepts, and yet it can be even more difficult for machines. 
The machine should have the ability of natural language processing and producing a human-understandable summary and background knowledge.
This is even more challenging, considering that different people have different interests and concerns to make their summaries, the subjectivity problem of summarization.
There exist different categorization for document summarization problem.
There exist different categories for document summarization.
For instance, one is based on the goal of the summarization task, which includes generic, domain-based (topic-focused)~\cite{codina2017using}, or query-based summarization algorithm~\cite{xiong2016query}. 
We also have other categories for document summarization which is based on the application of summarization such as article summarization~\cite{xu2019scientific}, review summarization~\cite{hu2017opinion}, news summarization~\cite{yang2019language}, and also summarization for anomaly detection~\cite{ahmed2019intelligent}.
In this paper, we consider the problem of summarization from a traditional perspective as well as recent personalized and interactive approaches, as discussed below.
\subsection{Traditional Approaches}
Traditional state-of-the-art approaches produce only a single, globally short summary for all users.
There are different perspectives to categorize traditional summarization approaches.
The main aspect is considering the process and the output type of the summarization algorithms, which include \textit{extractive} and \textit{abstractive approaches}. 
The problem in both tasks is defined as summarizing a set of related articles and producing a short (e.g., 3-6 sentences) single summary, which conveys the most informative information.
Abstractive approaches generate summaries by interpreting the main concepts of a document and then stating those contents in another format. 
Therefore, abstraction techniques are a substitute for the original documents rather than a part of it. 
Consequently, abstractive approaches require deep natural language processing, such as semantic representation and inference.
However, they are challenging to produce and yet have not arrived at a mature stage~\cite{gupta2010survey}. 
On the other hand, the extractive text summarization approach selects some sentences as representative of the original documents.
These sentences are then concatenated into a shorter text to produce a meaningful and coherent summary~\cite{heu2015fodosu}. 
Early extractive approaches focused on shallow features, employing graph structure, or extracting the semantically related words. 
Different machine learning approaches are also used for this purpose, such as naive-Bayes, decision trees, log-linear, and hidden Markov models~\cite{ghodratnama2009innovative,ghodratnama2015efficient,gupta2010survey}.
Recently, the focus for both extractive and abstractive approaches is mainly on neural network-based and deep reinforcement learning methods, which could demonstrate promising results.
They employ word embedding~\cite{pennington2014glove} to represent words at the input level.
Then, feed this information to the network to gain the output summary. 
These models mainly use a convolutional neural network~\cite{cao2015learning}, a recurrent neural network~\cite{cheng2016neural,nallapati2017summarunner} or combination of these two~\cite{narayan2018ranking,wu2018learning}. 
The problem is that these approaches do not consider the users' opinions and are not interactive. Consequently, the summaries are not well-tailored from the users' perspective.

\subsection{Personalized and Interactive Approaches}
While most state-of-the-art approaches produce a single general summary for all users, a few attempts to take a user's preferences into account are defined as personalized or interactive summarization techniques.

Interactive summarization approaches include approaches which require human to interact with the system to make summaries.
Unlike non-interactive systems that only present the system output to the end-user, Interactive NLP algorithms ask the user to provide certain feedback forms to refine the model and generate higher-quality outputs tailored to the user.
Most approaches in this category create a summary and then require humans to cut, paste, and reorganize the critical elements to make the final summary~\cite{narita2002web,orasan2006computer}.
Multiple forms of feedback also have been studied including mouse-clicks for information retrieval~\cite{borisov2018click}, post-edits and ratings for machine translation~\cite{denkowski2014learning}, error markings for semantic parsing~\cite{lawrence2018counterfactual}, and preferences for translation~\cite{kingma2014adam}. 
Other interactive summarization systems include the iNeATS~\cite{leuski2003ineats} and IDS~\cite{jones2002interactive} systems that allow users to tune several parameters (e.g., size, redundancy, focus) to customize the produced summaries.

The closest work to ours is proposed by Avinesh and Meyer~\cite{avinesh2017joint}, an interactive summarization approach that asks users to label important bigrams within candidate summaries.
Then they used integer linear programming (ILP) to extract sentences, covering as many important bigrams a possible. 
However, importance is a binary value in this system, important and unimportant.
The work by Orasan and Hasler~\cite{orasan2006computer} is also closely related to ours since they assist users in creating summaries for a source document based on the output of a given automatic summarization system. 
However, their system is neither interactive nor considers the user's feedback in any way.
Instead, they suggest the output of the state-of-the-art (single-document) summarization method as a summary draft and ask the user to construct the summary without further interaction.
The problem of concept-based ILP summarization framework was first introduced by~\cite{boudin2015concept}.
However, they used bigrams as concepts~\cite{berg2011jointly,li2013using} and either use document frequency (i.e. the number of source documents containing the concept) as weights~\cite{berg2011jointly,woodsend2012multiple}. 
As our interactive approach we allows for any combination of words, even sentence as concepts and also the weights are user defined parameters.

Our models also employ an optimization function to maximize user-desired content selection.
\textit{Adaptive Summaries} creates a personalized summary that better captures the users' needs and their different notions of importance by keeping the human in the loop.
Instead of binary labeling of concepts as important and unimportant, users can give feedback to either select or reject a concept, define the level of importance or being unrelated, and the user's level of confidence in providing an iterative feedback loop.
In the following section, we formalize the proposed approach.
\section{Adaptive Summaries} 
The goal of \textit{Adaptive Summary} is to interact with users to maximize the user-desired content in generating personalized summaries for users by interactions between system and user.
In this problem, the input is a set of documents where the output is a human-readable summary consisting of a set of sentences with the user's preferred size.
The novelty of this paper is that the user can select the desired content in making personalized summaries.
In this setting, users can choose either reject or accept action for selecting a concept being included in the summary, the importance of that concept from users' perspectives, and the confidence level of users' feedback.
This is modeled as an objective function to maximize the score of sentences based on the user-selected budget.
Besides, to guarantee interactive speed to keep the user engaged, we propose a heuristic approach for selecting users' queries.
In the following, we formally define the summarization tasks considered in this paper. 
\subsection{Problem Definition}
The input is a set of documents $\{D_{1},D_{2}, ... ,D_{N}\}$ while each document consists of a sequence of sentences $S=[s_1,s_2,$$...$$,s_n]$. 
Each sentence $s_i$ is a set of concepts $\{c_1,c_2, ..,c_k\}$ where a concept can be a word (unigram) or a sequence of words (Name entity or bigram).
This framework optimizes the summarization outcome for a specific user.
Therefore, the user interacts with the system and gives feedback to make summaries.
This feedback is in the form of:
i) Action $A$ which perform on a concept where the values can be \textit{accept (A=1)} or \textit{reject (A=-1)},
ii) concept weight, $W$, corresponding to concepts' importance according to the user's opinion, and
iii) the level of confidence for the chosen action, $ conf$.
The output is a set of sentences $S$ define as the summary according to the budget limit ($B$) defined by the user.

\subsection{Methodology}
The goal of \textit{Adaptive Summaries} is to incorporate the user preference in iteratively making summaries.
Therefore, a continuous objective function is defined for analytically optimizing the user preference.
In the first iteration, a summary is generated using our previous work, ExDos~\cite{samira2020}, that ranks sentences based on a general notion of importance using dynamic local feature weighting.
It also demonstrates sentences in groups based on similarity defined in ~\cite{samira2020} to help users in selecting content.
The user then can select an action $A$, which performs on a concept where the values can be \textit{accept (A=1)} or \textit{reject (A=-1)}.
Next, for each concept user can define a weight, $W$, corresponding to the concepts' importance based on the user's opinion. 
Next, the user defines the level of confidence, $conf$, for the chosen action.
When the action is accepted, this weight represents the importance of the concept, and when the action is rejection, the weights are the value for being unrelated.
The logic behind this is that not all concepts have an equal level of importance. 
For instance, when users search for an illness's symptoms, a \textit{headache} may not be as important as \textit{sneezing} from users' perspectives.
On the other hand, a \textit{fever} may not be as unrelated as \textit{acne}.
The overall objective function, which is an Integer Linear Programming(ILP), is defined as:

\begin{equation}
\label{eq1}
\begin{gathered}
    maximize \sum_{s_i\in D} \sum_{c_j\in s_i} A \times conf(A) \times W_{c_j}\\  
     s.t.  \hspace{1cm} \sum_{s\in Summary} length(s) <B 
\end{gathered}
\end{equation}

Where $A$ is the action, $c_j$ is the concept in a given sentence ($s_i$), $D$ the source documents, $W_{c_j}$ is the corresponding user-preference weight for the concept$c_j$ and $B$ is the summary length given by user.
The objective function~\ref{eq1} maximizes the occurrence of concepts with maximum weights and confident level.
The sudo-code of the proposed algorithm is reported in Algorithm~\ref{alg1}.
The following is the high-level description of our approach:
\begin{itemize}
    \item To accelerate the process of making a summary, in first iterations, the sentences are ranked by our previous approach, ExDos.
    Then these weights are updated based on users' feedback.
    \item In order to prevent users from being overwhelmed, the similar sentences using our previous approach, ExDos, are grouped and shown to the user simultaneously.
     \item If weights of a concept gets updated in an iteration, the weights are updated for every occurrence of that concept.
    \item If the user rejects a sentence ($A_{s_i}=-1$), then the weight of the sentence is set to zero ($W_{s_i}=0$). 
    However, the system does not update the weights of concepts included in the sentence as there may be different reasons for rejection of a sentence such as redundancy or not being important.
    \item A concept is only selected if it is present in at least one of the selected sentences.
    \item The number of sentences is a user parameter define in each iteration and the confidence in feedback is set 1 by default.
    \item If there are no more concepts to query, the process terminated.
    To optimize the summary creation based on user feedback, concept weights iteratively change the in the objective function. 
\end{itemize}


\begin{algorithm}
\SetAlgoLined
 \SetKwInOut{Input}{input}
 \SetKwInOut{Output}{output}
 \Input{Document Cluster D.}
 \Output{Optimal Summary Generated by user ($S$). }\
 Ranked Sentences $\leftarrow$ ExDos(D)\;
    \While{user is not satisfied}{
    Concepts $\leftarrow$ ExtractNewConcepts(Ranked Sentences)\;
        \If{Concepts $\neq$ $\emptyset$}{
            Ask user for action ($A$), importance($W$), and confidence($Conf$)\;
            Select sentences to maximize Equation~\ref{eq1}\;
        }
        return Summary(S)\;
    }
return Summary(S)\;
\caption{Adaptive Summaries}
\label{alg1}
\end{algorithm}

\section{Experiment}
In this section, we present the experimental setup for implementing and assessing our summarization model's performance.
We discuss the datasets, give implementation details, and explain how system output was evaluated.

\subsection{Data}
To compare the performance of \textit{Adaptive Summaries} with the existing leading approaches, experiments on two benchmark datasets, DUC2002\footnote{Produced by the National Institute of Standards and Technology (https://duc.nist.gov/)} and CNN/Daily Mail~\cite{hermann2015teaching}\footnote{ https://github.com/abisee/cnn-dailymail} are performed.
The documents are all from the news domain and are grouped into various topic clusters.
We analyze our system based on different criteria, including selecting different units of concepts, number of iterations, and the ROUGE score.

\subsection{Evaluation}
We evaluate the quality of summaries using ROUGE (Recall-Oriented Understudy for Gisting Evaluation) measure~\cite{lin2004rouge}\footnote{We run ROUGE 1.5.5: \url{http://www.berouge.com/Pages/defailt.aspx} with parameters -n 2 -m -u -c 95 -r 1000 -f A -p 0.5 -t 0} defined below.
It compares produced summaries against a set of reference summaries.
The three variants of ROUGE (ROUGE-1, ROUGE-2, and ROUGE-L) are used.
ROUGE-1 and ROUGE-2 are used to evaluate informativeness, and ROUGE-L (longest common subsequence) is used to assess the fluency. 
We used the limited length ROUGE recall-only evaluation (75 words) for comparison of DUC to avoid being biased.
Besides, the full-length F1 score is used for the evaluation of the CNN/DailyMail dataset.\newline
\begin{dmath}
ROUGE_n= \frac{\sum_{S\in \{Reference Summaries\}} \sum_{gram_n \in S}{Count_{match}(gram_n)}}{\sum_{S\in \{Reference Summaries\}} \sum_{gram_n \in S}{Count(gram_n)}}
\end{dmath}
In traditional approaches, to evaluate a summarization system, the mean ROUGE scores across clusters using all the reference summaries are reported.
\textit{Adaptive Summaries} is evaluated based on the mean ROUGE scores across clusters per reference summary in personalized summarization approaches.
It is worth mentioning that this approach aims at facilitating making summaries for individual users, not improving the general accuracy of summaries.
Since this approach is interactive, it requires humans to interact with the system for a user study based evaluation.
However, collecting data for different settings from different humans is too expensive.
Thus we simulate the users' behavior by generating feedback.
To simulate users' behaviors, we analyze two variations of the proposed approach. 
In the first approach (AdaptiveDictionary), to simulate the users' behavior, we define a dictionary for ten clusters of topics, including the essential concepts and weights with defined actions for each concept.
In the second one (AdaptiveReference), the reference summaries are considered as the users' feedback.
The concepts are essential if they are presented in the reference summary. 
Therefore, we assign the maximum weight for the presented concepts.
We compare our approach with both traditional and personalized approaches.
The results are reported in Table~\ref{tab:cnn} and Table~\ref{tab:duc2002} for both datasets.
\begin{table}[t]
\centering
\caption{ROUGE score comparison on CNN/DailyMail using F1 variant of ROUGE.}
\label{tab:cnn}
    \begin{tabular}{|l|c|c|c|}
        \hline
        Model & Rouge-1 Score & Rouge-2 Score & Rouge-L Score \\
        \hline
        LEAD-3 & 39.2 & 15.7 & 35.5  \\
         \hline
        NN-SE & 35.4 & 13.3 & 32.6\\
         \hline
        SummaRuNNer & 39.9 & 16.3 & 35.1 \\
         \hline
        HSSAS & 42.3 &  17.8 & 37.6 \\
         \hline
        BANDITSUM & 41.5 & 18.7 & 37.6 \\
         \hline
        \textbf{Adaptive dictionary} & 42.9 & 20.1  & 38.2 \\
        \hline
        \textbf{Adaptive reference}  &  41.4 & 19.7 &  32.1\\
        \hline
    \end{tabular}
\end{table}
\begin{table}[t]
    \centering
    \caption{ROUGE score (\%) comparison on DUC-2002 dataset.}\label{tab:duc2002}
    \begin{tabular}{|l|c|c|c|}
        \hline
             Model & Rouge-1 Score & Rouge-2 Score  & Rouge-L Score \\
        \hline
              LEAD-3 & 43.6 & 21.0 &N/A \\ 
         \hline
             NN-SE & 47.4 & 23.0 & N/A\\
         \hline
            \begin{tabular}{@{}c@{}}SummaRuNNer\end{tabular} & 46.6 & 23.1 & N/A \\
         \hline
            HSSAS & 52.1 & 24.5 & N/A\\
         \hline
            Upper Bound & 47.4 & 21.6 &  18.7\\
         \hline
            Avinesh-Al & 44.8 & 18.8 &  16.8\\
         \hline
            Avinesh-Joint & 44.4 & 18.2 &  16.5\\
         \hline
            \textbf{Adaptive dictionary} & 50.4 & 22.1 & 18.4 \\
         \hline
            \textbf{Adaptive reference}  & 46.5  & 20.1 & 18.8\\
        \hline
    \end{tabular}
\end{table}
From the results, it can be seen that the proposed approach nearly reaches the upper bound for both datasets.
Besides, the ROUGE analysis with real users does not show any pattern of increasing or decreasing.
However, it is an expected result since this approach aims to optimize the summary for individual users, not the gold standard summary.

\begin{figure*}[t]
    \centerline{\includegraphics[width=1.2\textwidth]{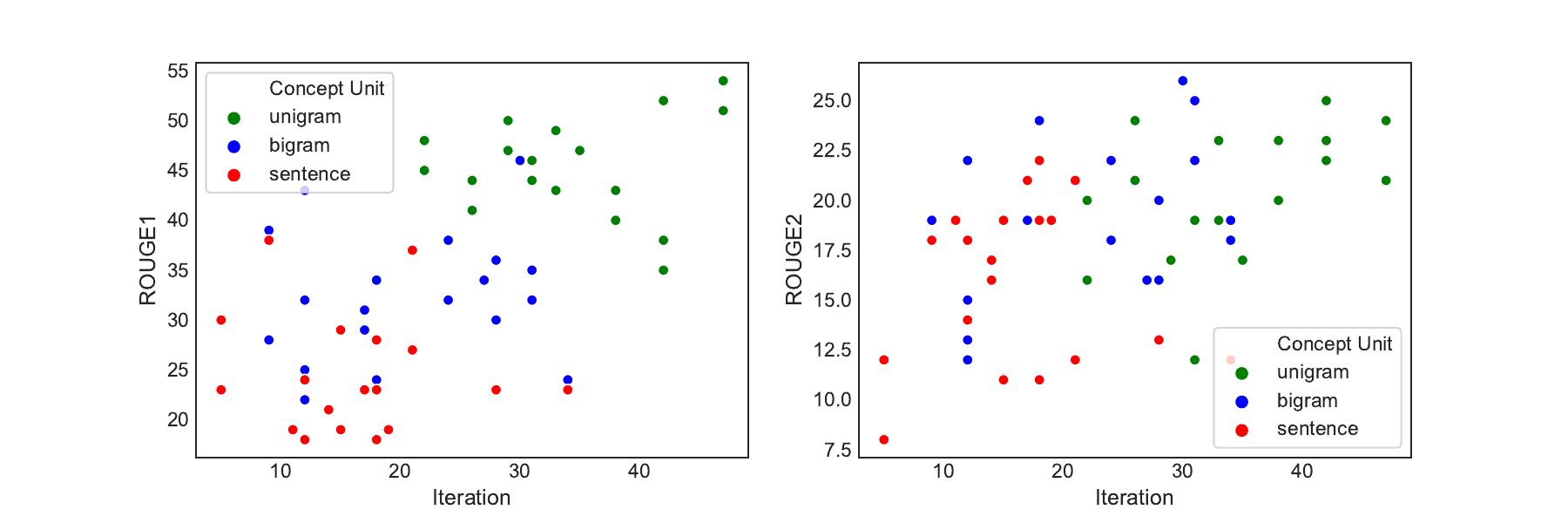}}
    \caption[]{Th left image shows the ROUGE1 based on iteration number and the right image shows the ROUGE2 based on iteration number. The green samples are when the permitted concept unit are unigram, blue bigram and red the sentences.}
    \label{conceptunit}
\end{figure*}

To compare the concepts' unit's effect, we evaluate our approach based on three-unit measures, including uni-gram, bi-gram, and sentences.
Although our model reaches the upper bound when using unigram-based feedback, they require significantly more iterations and much feedback to converge, as shown in Figure~\ref{conceptunit}.
We analyze the speed (iterations) and the accuracy (ROUGE1 and ROUGE2) for different concepts units for DUC2002.
CNN/DailyMail dataset follows the same trend.
From the image, we see that when the permitted selection unit is unigram, the ROUGE1 score is higher. However, it takes more iterations to converge.
For ROUGE2, both bigram and unigram have higher scores, however, when the unit is bigram, it converges sooner.

\begin{figure*}[t]
    \centerline{\includegraphics[width=1.2\textwidth]{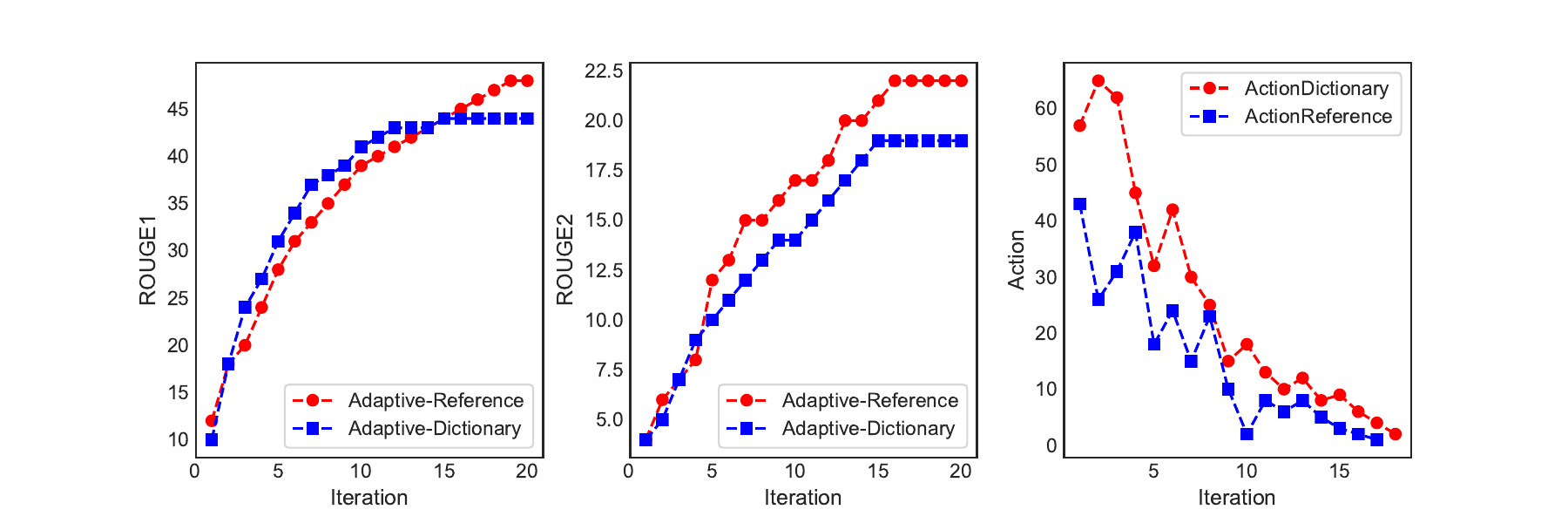}}
    \caption[]{1) Number of iteration and ROUGE1 for DUC-2002. 2) Number of iterations versus ROUGE2 values. 3) Number of actions versus iterations for DUC2002.}
    \label{iterations}
\end{figure*}

Another experiment is considering the ROUGE scores versus the number of iterations.
In Figure~\ref{iterations}, the results for the DUC-2002 data set for two versions of adaptive, using a dictionary as feedback and reference summary as feedback is depicted.
In the third image, we evaluated the models based on the number of actions (A) taken by the oracles to converge to the upper bound within ten iterations.


\section{Conclusion and Future Work} 
We propose an interactive and personalized multi-document summarization approach using users' feedback.
The selection or rejection of concepts, defining the importance of a concept, and the level of confidence engage users in making their desired summary.
We empirically checked the validity of our approach on standard datasets using simulated user feedback.
We observed that our framework shows promising results in terms of ROUGE score and also human evaluation.
Results show that users' feedback can help them to find their desired information.
As future work, we plan to include the reasons behind any action to optimize the system's performance.

\textbf{\emph{Acknowledgement.}}
We acknowledge the AI-enabled Processes (AIP) Research Centre  \footnote{https://aip-research-center.github.io/} for funding this research.
We also acknowledge Macquarie University for funding this project through IMQRES scholarship.

\bibliographystyle{splncs04}
\bibliography{Main.bib}
\end{document}